\documentclass[10pt,twocolumn,letterpaper]{article}

\usepackage{wacv}
\usepackage{times}
\usepackage{epsfig}
\usepackage{graphicx}
\usepackage{amsmath}
\usepackage{amssymb}
\usepackage{booktabs}
\usepackage[accsupp]{axessibility}  %
\makeatletter
\@namedef{ver@everyshi.sty}{}
\makeatother
\usepackage{tikz}
\usepackage{comment}
\usepackage{amsmath,amssymb,bbm} %
\usepackage{color}
\usepackage{multirow}
\usepackage[normalem]{ulem}
\usepackage{booktabs}
\input{def}

\def\wacvPaperID{772} %

\wacvalgorithmstrack   %

\wacvfinalcopy %

\ifwacvfinal
\usepackage[pagebackref=true,breaklinks=true,bookmarks=false]{hyperref}
\else
\usepackage[pagebackref=true,breaklinks=true,colorlinks,bookmarks=false]{hyperref}
\fi

\begin{document}

\title{Towards Disturbance-Free Visual Mobile Manipulation}

\author{
Tianwei Ni$^{1,2,*}$ \quad  Kiana Ehsani$^{2}$ \quad Luca Weihs$^{2,\dagger}$ \quad Jordi Salvador$^{2,\dagger}$  \\
{\normalsize $^1$Université de Montréal \& Mila -- Quebec AI Institute}  \quad {\normalsize $^2$PRIOR @ Allen Institute for AI}
\\
{\footnotesize \texttt{tianwei.ni@mila.quebec} \quad \, \texttt{$\{$kianae, lucaw, jordis$\}$@allenai.org}}
\\
{\footnotesize  $^*$ Work was primarily done during internship at AI2 \quad $\dagger$ Equal advising}
}

\maketitle

\begin{figure*}
    \centering
    \includegraphics[width=0.9\linewidth]{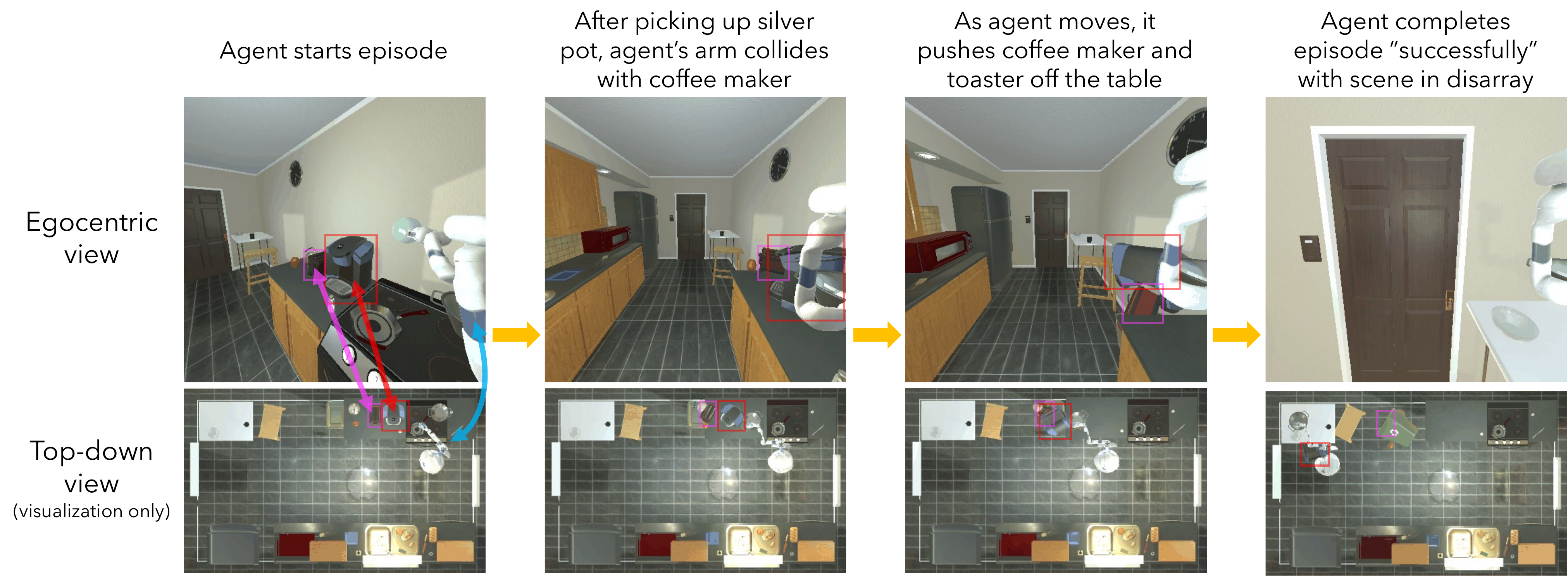}
    \caption{\textbf{Success \emph{with} disturbance.}
    A robotic ManipulaTHOR~\cite{ehsani2021manipulathor} agent attempts to complete an ArmPointNav task where it must pick up and move one object from one side of a kitchen to another. A well-trained agent (in white) successfully moves the target object (a silver pot) from a source location (stove burner) to a target location (dining table). Despite this success, the agent disturbs several objects (including a \textcolor{red}{coffee maker} and a \textcolor{magenta}{toaster})
     when moving its arm and body. Both the coffee maker and toaster are pushed off the countertop and the coffee maker is further pushed across the kitchen, a catastrophic outcome in the real world. 
    The figures show both RGB egocentric and top-down views, while the agent only has access to egocentric observations.}
    \label{fig:teasor}
    \vspace{-1em}
\end{figure*}

\thispagestyle{empty}

\begin{abstract}

Deep reinforcement learning has shown promising results on an abundance of robotic tasks in simulation, including visual navigation and manipulation. Prior work generally aims to build embodied agents that solve their assigned tasks as quickly as possible, while largely ignoring the problems caused by collision with objects during interaction. This lack of prioritization is understandable: there is no inherent cost in breaking virtual objects. As a result, ``well-trained'' agents frequently collide with objects before achieving their primary goals, a behavior that would be catastrophic in the real world. In this paper, we study the problem of training agents to complete the task of visual mobile manipulation in the ManipulaTHOR environment while avoiding unnecessary collision (disturbance) with objects. We formulate disturbance avoidance as a penalty term in the reward function, but find that directly training with such penalized rewards often results in agents being unable to escape poor local optima. Instead, we propose a two-stage training curriculum where an agent is first allowed to freely explore and build basic competencies without penalization, after which a disturbance penalty is introduced to refine the agent's behavior. Results on testing scenes show that our curriculum not only avoids these poor local optima, but also leads to 10\% absolute gains in \textit{success rate without disturbance}, compared to our state-of-the-art baselines. Moreover, our curriculum is significantly more performant than a safe RL algorithm that casts collision avoidance as a constraint. Finally, we propose a novel disturbance-prediction auxiliary task that accelerates learning.
\footnote{Project page is at \url{https://sites.google.com/view/disturb-free}}

\end{abstract}

\vspace{-3mm}
\section{Introduction}

Advances in deep reinforcement learning (RL) for embodied agents has led to significant progress in visual navigation~\cite{mirowski2016learning,zhu2017target,kolve2017ai2,xia2018gibson,savva2019habitat,wijmans2019dd,chaplot2020neural,ye2020auxiliary} and manipulation~\cite{kalashnikov2018qt,zeng2018learning,gualtieri2018learning,wu2020spatial,james2020rlbench,gan2020threedworld,puig2018virtualhome}. 
In this paper, we focus on the relatively new embodied-AI problem of \textbf{visual mobile manipulation}~\cite{li2017reinforcement,shen2020igibson,xiang2020sapien}, in particular the ArmPointNav task proposed by Ehsani \etal~\cite{ehsani2021manipulathor} set in the simulated mobile manipulation framework ManipulaTHOR. In ArmPointNav, the goal is to bring an object to a goal location. An agent must perform navigation and manipulation jointly by navigating to an object of interest, picking the object up using its attached 6-DOF robotic arm, and then carrying the object to a target location.
Combining navigation and manipulation, especially when agents are expected to generalize to novel scenes and objects, 
is a challenging but important step towards building generally capable household robotic agents.
Similarly, as for other embodied AI tasks, RL methods attain respectable performance for ArmPointNav, with a baseline achieving a success rate of 62\%~\cite{ehsani2021manipulathor}.

These relatively high success rates for ArmPointNav come, however, with a significant caveat: success only requires that an agent manages to bring an object to a goal and entirely ignores whether the agent collides with other objects during interaction. Ignoring \textbf{collision avoidance} when measuring success is not exclusive to ArmPointNav, but ubiquitous in pure navigation~\cite{wijmans2019dd,ye2020auxiliary,deitke2020robothor}, 
pure manipulation~\cite{levine2018learning,kalashnikov2018qt}, and mobile manipulation~\cite{sun2021relmm,wang2020learning} tasks, even despite there being collision detection capabilities in relevant simulators~\cite{szot2021habitat,shen2020igibson}.
Furthermore, the popular SPL metric (success weighed by path length)~\cite{anderson2018evaluation} may even encourage collision, as it rewards agents for taking shortcuts that may result in collisions. 
With the introduction of mobile manipulation, where agents have more chances to interact with objects, the catastrophic impact of collisions becomes too obvious to ignore: 
Fig.~\ref{fig:teasor} shows a prototypical example where an agent ``successfully'' completes the ArmPointNav task, but disturbs many objects in the scene.
In fact, the ArmPointNav baseline of Ehsani \etal has only a 30\% success rate \emph{without disturbance} (collision) with other objects. 
This means that even a (relatively) high success rate should give little confidence that a policy can be safely deployed in the real world.

Unlike in the robotics community, where the importance of safety and collision avoidance are deeply ingrained~\cite{fox1997dynamic,lavalle2001randomized,perkins2002lyapunov}, the incentives for ignoring collisions in simulated embodied AI tasks are clear: (1) there is no inherent cost of collision in \emph{simulation} and (2) accurate collision detectors are computationally expensive and so, as speed is one of the great advantages of simulation, environments frequently simplify or ignore the impact of collisions altogether~\cite{savva2019habitat,kolve2017ai2}.
As algorithms improve and task success rates increase, this practice of ignoring collisions will prevent real world deployment where
the costs of breaking objects, damaging robots, and harming humans are unacceptable~\cite{flacco2012depth}.

In this paper, towards enabling safe deployment of real robots, we propose to train embodied agents while prioritizing what we call \textbf{disturbance avoidance}: namely, we require that agents avoid moving any object that is not directly related to the agents' goal. For flexibility, we only consider object positions at the start and end of an episode. For example, this allows an agent to temporarily move an object that would otherwise prevent it from reaching its goal.
While collision avoidance and disturbance avoidance may, at first, seem synonymous, the problem of collision avoidance is strictly more difficult than disturbance avoidance. To see this, note that not all collisions need result in the positions of objects being changed. For instance, an agent might run into a wall (a collision) but, as the wall is not moved by this collision, the scene is undisturbed.
Indeed, disturbance avoidance can be thought of as ``visible collision avoidance'':  collisions that do not result in a visual change in the environment are ignored.
We study disturbance avoidance, instead of collision avoidance, for three reasons: (1) disturbance can be efficiently measured in simulation as it requires only checking how the pose of objects has changed between time-steps, while collision detection requires measuring contact forces; (2) our agents take as input purely egocentric visual inputs and thus have a limited capacity to learn about phenomena that they cannot directly observe; (3) disturbance avoidance allows for the temporary movement of the objects so long as they are eventually moved back to their original positions, a practical necessity for some tasks.

A standard approach for encouraging safe behavior in RL agents is simply to penalize unsafe behavior~\cite{kahn2017uncertainty,richter2017safe,tai2017virtual,li2020hrl4in}; in this vein, we modify the standard ArmPointNav reward structure by introducing a penalty for object disturbance.
Perhaps surprisingly, in practice, training agents \emph{from scratch} using this new reward structure results in extremely unstable learning. Indeed, in many cases, agents trained with this reward structure learn not to disturb any objects by terminating early without reaching the goal: a bad local optimum. We hypothesize that this object disturbance penalty discourages early exploration and thus results in a highly suboptimal policy.

Inspired by this empirical finding, we propose a simple but effective \textbf{two-stage training curriculum}: we first train the agent with original reward (without penalty), and then fine-tune the agent with penalized reward. During the first stage, the agent can learn to solve the task \textit{with} disturbance as it has enough freedom to explore, while in the second stage the agent can learn to adjust its behavior to avoid disturbance when solving the task.

In what follows, we first focus on the original ArmPointNav task with its original reward and success criteria. For this task, we show how several critical design decisions result in dramatic improvements over the existing state-of-the-art~\cite{ehsani2021manipulathor}: an 11.1\% absolute increase in success rate (SR) on testing scenes with novel objects (same evaluation setting below) after 20M training frames. 
With 45M training frames, our improved model further attains a success rate of 82.7\% but, critically, only achieves a 35.5\% success rate without disturbance (SRwoD). 
We then move on to our main focus: disturbance avoidance in ArmPointNav. We find that with the same 45M frames budget, our two-stage training is much more effective than training from scratch, with 80.1\% vs 18.0\% in SR, and 46.5\% vs 10.5\% in SRwoD. In other words, two-stage training avoids performance degradation and achieves higher SRwoD than our improved baseline using the original objective (46.5\% vs 35.5\%).
It also outperforms PPO-Lagrangian~\cite{ray2019benchmarking}, a popular safe RL algorithm, by over 30\% in SRwoD. 
Finally, we propose a new supervised auxiliary task that requires the agent to predict how its actions will disturb the environment and show that co-training with the auxiliary task can accelerate learning and increase final performance when compared to using no (or self-supervised) auxiliary tasks.

In summary, in this paper we present the following contributions: 
(1) we propose a disturbance avoidance objective for embodied RL agents, 
(2) we introduce a state-of-the-art model for the original ArmPointNav task with extensive ablative experiments, 
(3) we provide strong empirical evidence that our two-stage curriculum can lead to agents which avoid disturbance while retaining high success rates, and 
(4) we offer a new auxiliary disturbance-prediction task which accelerates learning.

\section{Related Work}

\begin{table}[t]
    \centering
    \footnotesize
    \begin{tabular}{c|cccc}
    \toprule 
    \multirow{2}{*}{Methods} & Navi- & Manipu- & Avoid & \multirow{2}{*}{Framework} \\
    & gation? & lation? & Collision? & \\
     \midrule
     \cite{zhu2017target} & \cmark & \xmark & \xmark & MDP \\ 
     \cite{gupta2017cognitive,mirowski2016learning,wijmans2019dd,ye2020auxiliary} & \cmark & \xmark & \xmark & POMDP \\
     \cite{richter2017safe} & \cmark & \xmark & \cmark & Supervised \\ 
     \cite{kahn2018self,kahn2021badgr} & \cmark & \xmark & \cmark & MDP \\ 
     \cite{pinto2016supersizing,viereck2017learning,morrison2018closing,zeng2018robotic} & \xmark & \cmark & \xmark  & Supervised \\ 
     \cite{mousavian20196,murali20206}  & \xmark & \cmark & \cmark  & Supervised \\ 
     \cite{levine2018learning,kalashnikov2018qt,zeng2018learning,gualtieri2018learning,song2020grasping} & \xmark & \cmark & \xmark  & MDP \\
     \cite{mahler2017learning} & \xmark & \cmark & \cmark  &  POMDP \\ 
     \cite{li2020hrl4in,wu2020spatial}  & \cmark & \cmark & \cmark  & MDP \\
     \cite{wang2020learning,sun2021relmm} & \cmark & \cmark & \xmark  & MDP \\ 
     \cite{ehsani2021manipulathor} & \cmark & \cmark & \xmark  & POMDP \\ 
     Ours & \cmark & \cmark & \cmark  & POMDP \\ 
 
     \bottomrule
    \end{tabular}
    \vspace{2mm}
    \caption{\textbf{Data-driven methods for visual navigation and manipulation.} We classify the methods by the studied problem (navigation, manipulation, or both, \ie, mobile manipulation), whether they \textbf{learn to} avoid collision (disturbance), and their adopted algorithmic frameworks (supervised learning, MDP~\cite{bellman1957markovian}, and POMDP~\cite{aastrom1965optimal}).}
    \vspace{-2em}
    \label{tab:related_work}
\end{table}

\noindent\textbf{Visual navigation and manipulation.}
A long history of work exists for both (stationary) tabletop manipulation using fixed-based arms~\cite{khatib1995inertial,rahardja1996vision,jin2016multi}, pure embodied navigation~\cite{mirowski2016learning,zhu2017target},
and mobile manipulation~\cite{yamamoto1992coordinating,khatib1999mobile,wolfe2010combined,hornung2012navigation,stulp2012learning,mittal2021articulated,li2017reinforcement}.
Much of this prior work, especially traditional methods for manipulation~\cite{schulman2013finding,prattichizzo2016grasping,berenson2007grasp,khatib1986real,quinlan1993elastic,ratliff2009chomp,brock1998mobile,dogar2011framework,miller2003automatic,kaelbling2012unifying,stilman2005navigation}, requires ground-truth knowledge of objects and the environment (such as their geometry) and becomes computationally expensive in high-dimensional settings. 
Data-driven methods relax these assumptions and enable agents to act from visual inputs.
We summarized the data-driven related work in Table~\ref{tab:related_work} from several perspectives.
Our work focuses on ArmPointNav~\cite{ehsani2021manipulathor}, a \emph{visual} mobile manipulation task where agents must navigate in the ManipulaTHOR environment to find an object, pick it up using an attached arm, and then bring the object to a new location. The ManipulaTHOR environment consists of a set of visually complex scenes, supports navigation with a mobile robot, and allows for object manipulation with a 6-DOF arm through clutter.
ArmPointNav follows the more general POMDP framework that only has egocentric depth observations and 3D goal sensors without other state information. 
Our method is built upon a recurrent model-free RL baseline provided by ManipulaTHOR, and focuses on disturbance avoidance discussed below.

\noindent\textbf{Collision/disturbance avoidance.}
In safety-critical domains, collision avoidance is extremely important.
Classic methods in motion planning~\cite{fox1997dynamic,lavalle2001randomized,perkins2002lyapunov,majumdar2017funnel,daftry2016introspective} and path planning~\cite{khatib1986real,hwang1992potential,cherubini2011visual}, which provide safety guarantees, require privileged information of obstacles to avoid collision and are hard to scale to partially-observable visually complex settings. 
Data-driven methods learn to avoid collision from data with less privileged information (see the ``collision" column in Table~\ref{tab:related_work} for a summary).
Deep-RL methods, which can learn collision avoidance from interaction within environments, can be divided into model-free and model-based approaches. 
Model-free methods simply introduce collision penalties in the reward function, in the form of constants when facing collision~\cite{tai2017virtual,long2018towards,wu2020spatial,li2020hrl4in,feng2021collision} or being proportional to the distances to the nearest obstacles~\cite{kahn2017plato,sangiovanni2018deep,everett2021collision,kahn2021badgr}. 
Model-based methods~\cite{kahn2017uncertainty,kahn2018self,kahn2021badgr} explicitly learn a collision prediction model and use it for policy search. 
Our work studies disturbance, a subset of collision, where an object is moved by some distance, because disturbance is easier to compute in simulation and allows temporary displacement. 
We first consider a \textit{model-free} setting (adding a disturbance penalty) and show that it can perform well when paired with our two-stage training curriculum. We then add a \textit{model-based} component (a disturbance prediction auxiliary task) to accelerate learning.

\noindent\textbf{Constrained MDPs and safe RL.} An alternative formulation of collision avoidance is to frame avoidance as \textit{constraint} (not a fixed penalty in reward) during policy optimization, \ie, frame the problem as a constrained MDP~\cite{altman1999constrained}. Algorithms solving constrained MDPs are frequently employed in the domain of safe RL. A popular approach in this area is to employ Lagrange multipliers in various RL algorithms~\cite{ray2019benchmarking,liang2018accelerated,tessler2018reward,ha2020learning} allowing for the adaptive penalization of unwanted behavior. Despite being adaptive and agnostic to the reward scale, Lagrangian methods have been shown to be sensitive to the initialization of the multipliers~\cite{achiam2017constrained}. In our experiments, we find our approach to be more performant than a competing, Lagrangian-based, safe RL baseline.

\noindent\textbf{Transfer and curriculum learning in RL.} Transfer learning is widely used in deep learning to transfer knowledge from a source domain to a target domain~\cite{hinton2006reducing,girshick2014rich,yosinski2014transferable}.
Transfer learning is also popular in RL~\cite{kalashnikov2018qt}, such as continual RL setting~\cite{rusu2016progressive,julian2020never}.
Our two-stage training methodology reframes the single task learning (success without disturbance) into ``curriculum learning''~\cite{bengio2009curriculum,narvekar2020curriculum} on a task sequence, where the first task (stage) is to succeed with disturbance, and the second task (stage) is to succeed without disturbance. 
We show that this curriculum formulation is far more effective than direct learning on the final (hard) task, as learning on the early task is much easier and bootstraps learning the final task. 

\noindent\textbf{Auxiliary tasks in RL.} Auxiliary tasks, co-trained with the main task (maximizing the total rewards) on the shared model weights, have been shown to have the promising capability to improve sample efficiency and asymptotic performance in visual RL.
Supervised auxiliary tasks provide extra information to the policy via external signals, such as depth maps~\cite{mirowski2016learning,das2018embodied} and game internal states~\cite{lample2017playing}, 
while self-supervised/unsupervised auxiliary tasks use existing information as signals, such as auto-encoders~\cite{lange2010deep,ha2018recurrent,yarats2019improving}, forward~\cite{gregor2019shaping} and inverse dynamics~\cite{pathak2017curiosity}, reward prediction~\cite{jaderberg2016reinforcement}, and contrastive learning~\cite{guo2018neural,guo2020bootstrap,ye2020auxiliary,srinivas2020curl}. 
Our work introduces a supervised auxiliary task that predicts the ``disturbance distance'', namely, how greatly an agent's action will disturb the environment. Since this disturbance distance is one component of our reward, this task can be viewed as distilling knowledge of the reward's composition to the agent.

\section{Towards Disturbance-Free ArmPointNav}
\label{sec:method}
As preliminaries, we first introduce our model architecture (Sec.~\ref{sec:baseline}) for ArmPointNav.
Then we formulate the concept of disturbance avoidance along with a new, corresponding, reward structure for ArmPointNav (Sec.~\ref{sec:penalty}).
Next, we define a new auxiliary task, disturbance prediction, which could improve sample efficiency in training (Sec.~\ref{sec:task}).
Finally, we discuss the training techniques on the new objective, and introduce our two-stage training curriculum (Sec.~\ref{sec:curriculum}) that is most critical to final performance.

\begin{figure*}
    \centering
    \includegraphics[width=0.9\linewidth]{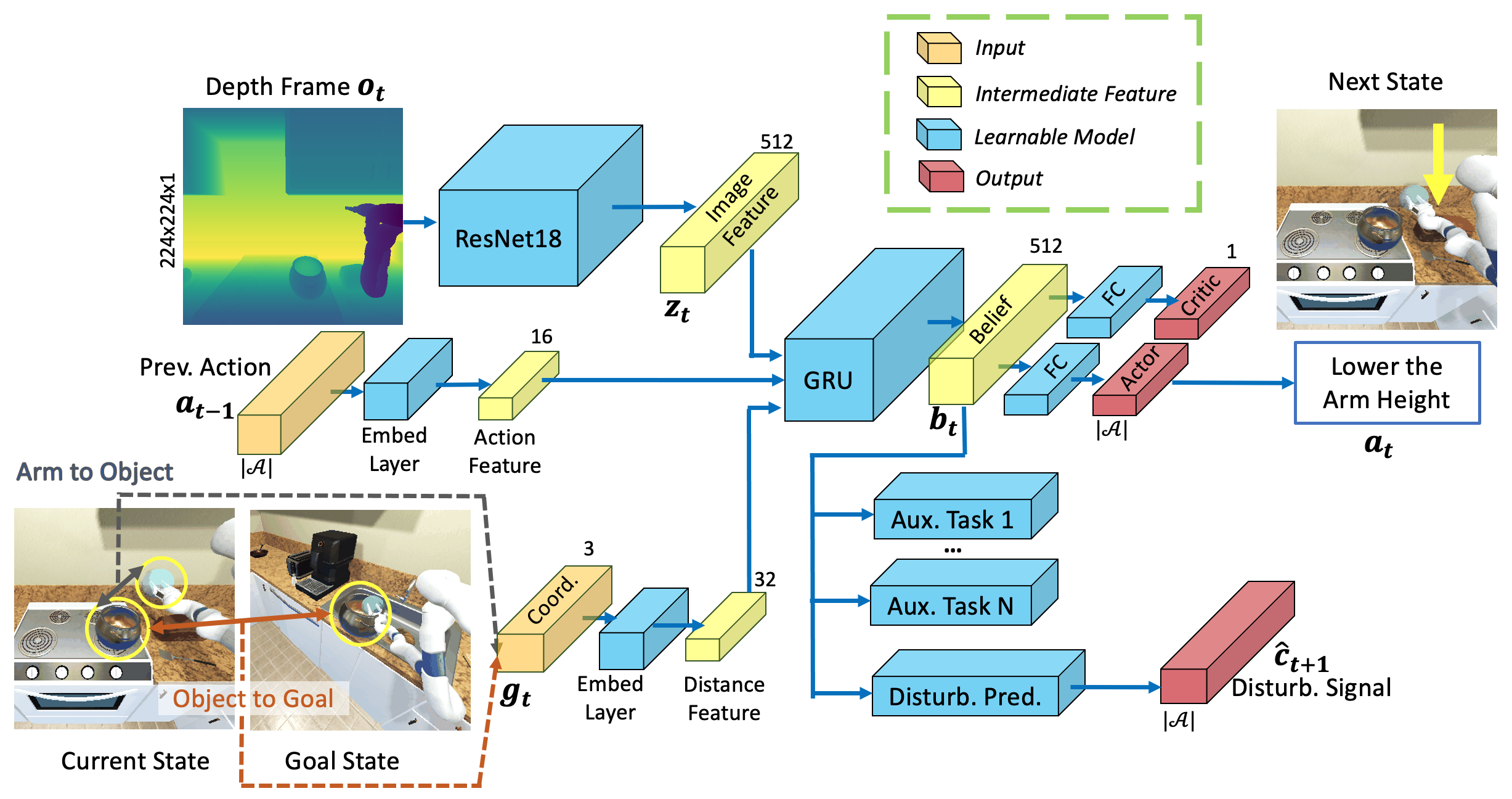}
    \vspace{-1mm}
    \caption{\textbf{Our model architecture for ArmPointNav.} We make several improvements to the existing SoTA baseline from \cite{ehsani2021manipulathor}. 
    These improvements include: replacing the existing shallow CNN with a ResNet18, adding an embedding of the previous action $a_{t-1}\in \mathcal A$ to the agent's inputs, and using polar, instead of Cartesian, goal coordinates $g_t \in \mathbb R^3$. These changes greatly increase performance (see Sec.~\ref{sec:exp_baselines}).
    Building on the work of Ye \etal~\cite{ye2020auxiliary}, we also enable support for training with arbitrary self-supervised auxiliary tasks given current belief $b_t$ for sample efficiency. 
    Finally, we add a specific auxiliary task head for our novel next-step disturbance prediction task $\hat c_{t+1}\in {\mathbb R}^{|\mathcal A|}$.}
    \label{fig:arch}
    \vspace{-6mm}
\end{figure*}

\subsection{Preliminaries: Model Architecture}
\label{sec:baseline}

For the task of ArmPointNav, given only egocentric depth observations and goal coordinates, an agent must interact with its environment so as to to pick up a target object with its arm and then navigate to place that target object in a goal location. 
See App.~\ref{appendix:env} for more details.

We now describe our model architecture for ArmPointNav, based on the original baseline proposed by Ehsani \etal~\cite{ehsani2021manipulathor}. At a timestep $t\geq 0$, the model takes as input the current egocentric depth observation $o_t$, which is passed through a modified ResNet18~\cite{he2016deep,wijmans2019dd} to produce the embedding $z_t = \mathrm{ResNet}(o_t)$. The model then passes this embedding $z_t$ along with an embedding of distance to goal $g_t$, the encoding of previous action $a_{t-1}$, and the belief state $b_{t-1}$ from the previous timestep, to a single-layer GRU~\cite{chung2014empirical} to produce the current belief state  $b_t = \mathrm{GRU}(b_{t-1}, \mathrm{ResNet}(o_t), g_t, a_{t-1})$.
Finally, following an actor-critic formulation, a linear layer is applied to the belief state $b_t$ to produce the agent's policy $\pi(b_t)$ (i.e, a distribution over actions) and an estimate of the value of the agent's current state $V(b_t)$. 

Fig.~\ref{fig:arch} shows a summary of this architecture, and the new design choices we made compared to the original baseline.
This agent is trained using the PPO algorithm~\cite{schulman2017proximal,wijmans2019dd} to maximize the discounted cumulative rewards $\sum_{t=0}^T \gamma^t r_t$, with $\gamma=0.99$ and $T=200$. The reward function $r_t$ is defined in Eq.~1 in~\cite{ehsani2021manipulathor}.

\vspace{-0.5em}
\subsection{Formulation of Disturbance Avoidance}
\label{sec:penalty}

Now we focus on the goal of training agents that are able to avoid disturbance.
One of the simplest strategies for discouraging unwanted behavior in RL is to incorporate a penalty into the reward function for this behavior.
We consequently add a penalty to the original reward function $r_t$ (Eq.~1 in~\cite{ehsani2021manipulathor}) to define a new reward function $r'_t$:
\begin{equation}
\label{eq:reward}
r'_t = r_t + \lambda_{\mathrm{disturb}} (d_{t-1}^{\mathrm{objects}} - d_{t}^{\mathrm{objects}})\ ,
\end{equation}
where $\lambda_{\mathrm{disturb}} > 0$ is a coefficient controlling the magnitude of the disturbance penalty, and $d_{t}^{\mathrm{objects}}$ is the sum of 3D Euclidean distances of all objects (except the target object) from their initial locations at time $t$.

The \textbf{disturbance-free objective} is now defined as the discounted cumulative sum of these new rewards:
\begin{equation}
\label{eq:disturb_free_obj}
\sum\nolimits_{t=0}^T \gamma^t r'_t = \sum\nolimits_{t=0}^T \gamma^t r_t + \gamma^t \lambda_{\mathrm{disturb}} (d_{t-1}^{\mathrm{objects}} - d_{t}^{\mathrm{objects}})\ .
\end{equation}
Notice that if $\gamma=1$ then the above sum telescopes to simply equal $(\sum_{t=0}^T r_t) - \lambda_{\mathrm{disturb}}d_{T}^{\mathrm{objects}}$.
Thus, up to discounting, the disturbance-free objective can be interpreted as the original ArmPointNav objective with a soft constraint on the final total disturbance distance.
This new objective discourages the agent from ending the episode with objects out of their original positions, but allows the agent to temporarily move objects as long as the agent eventually moves them back into their original locations. This behavior emerges in training (see Sec.~\ref{sec:exp_main}).

\subsection{Disturbance Prediction as an Auxiliary Task}
\label{sec:task}

Beyond passing knowledge indirectly via a model-free method (\ie, with disturbance penalty as in Eq.~\ref{eq:disturb_free_obj}), 
to improve sample efficiency, we also consider a model-based approach that explicitly requires the agent to predict object disturbance as an auxiliary task. 
Formally, given its current belief $b_t$ about the environment and the action $a_t$ it has decided to enact, the agent must predict the probability $\hat{c}_{t+1}$ of current action disturbing the environment. To produce this probability estimate in practice, we use an MLP denoted as $\mathtt{Disturb}$: 
\begin{equation}
\label{eq:disturb}
\mathtt{Disturb}(b_t, a_t) = \hat c_{t+1} \in [0,1]\ .
\end{equation}

We obtain the ground-truth \textit{binary} disturbance signal $c_{t+1}$ from the simulated environment: it is calculated by thresholding the current change in the disturbance distance:
\begin{equation}
c_{t+1}=\mathbbm{1}\left(d_{t+1}^{\mathrm{objects}} - d_{t}^{\mathrm{objects}}\ge \tau\right) \in \{0,1\}\ ,
\end{equation}
where $\tau\approx 0.001$~m in our case.

As most (${\approx}$90\%) actions taken by the agent during training do not result in disturbance, there is a significant class imbalance for this task formulation. To mitigate this imbalance, we leverage the Focal loss~\cite{lin2017focal}, a modified cross entropy loss designed for class-imbalanced prediction.
For every step $t$ in training, we compute the total loss for our agent as the sum of the usual actor, critic, and entropy losses from PPO as well as the auxiliary disturbance prediction loss $\mathcal{L}_{\mathrm{Focal}}(\hat  c_t, c_t)$. 

Our auxiliary task works in concert with our new objective: our auxiliary task can directly teach the agent to recognize the types of actions that result in disturbance, while our new objective encourages the agent to avoid such actions.

\subsection{Two-Stage Training Curriculum}
\label{sec:curriculum}

\begin{figure}[h]
    \centering
    \includegraphics[width=0.7\linewidth]{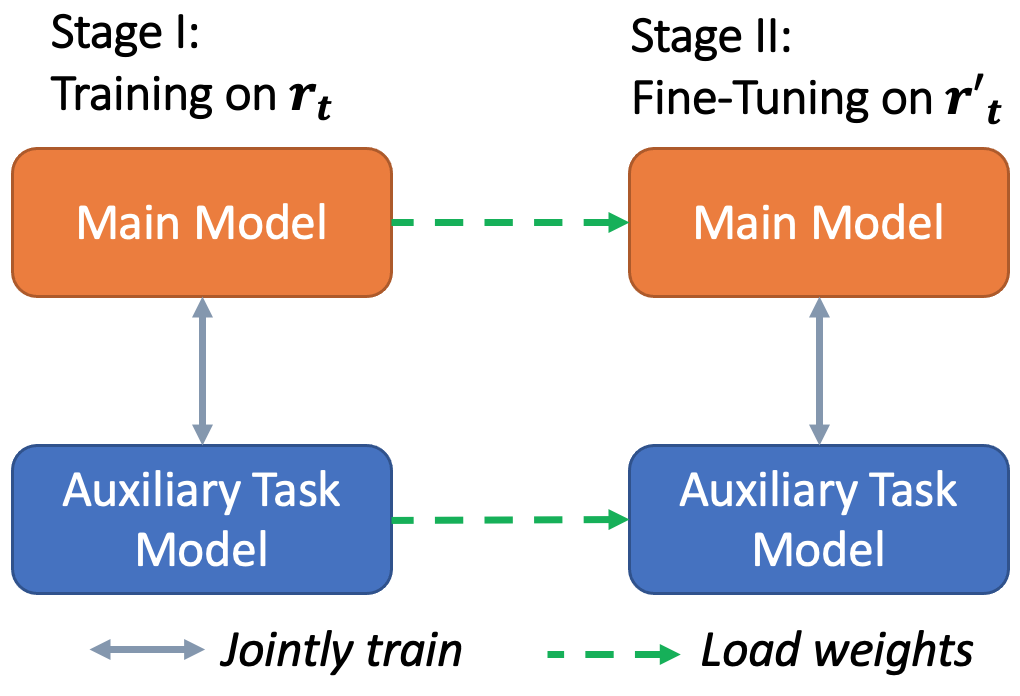}
    \vspace{-1mm}
    \caption{\textbf{Our two-stage training curriculum.} $r_t$ and $r'_t$ (Eq.~\ref{eq:reward}) are the original and new reward function, respectively. The main model refers to the model components other than those of auxiliary tasks. Co-training with auxiliary task is optional.}
    \label{fig:curriculum}
    \vspace{-1em}
\end{figure}

We train an agent using RL on the disturbance-free objective (Eq.~\ref{eq:disturb_free_obj}), with an option of co-training any auxiliary task (\eg, Sec.~\ref{sec:task}). 
Directly training to optimize for the new objective \textit{from scratch} is straightforward but, in practice, suffers from extreme instability with some ``trained'' models achieving near-zero success rates (see Sec.~\ref{sec:exp_main}). 
As the disturbance penalty discourages the agent from interacting with the objects, we hypothesize that it may also prevent the agent from sufficiently exploring potential strategies before settling on a conservative approach: giving up partway through an episode to avoid potential object disturbance.

Inspired by the empirical findings and work in curriculum learning, we propose a two-staged training curriculum for learning with the disturbance-free objective (see Fig.~\ref{fig:curriculum}).  
In the first stage, the agent is trained with the original reward function $r_t$ (potentially co-trained with auxiliary tasks). As the agent is not penalized for causing disturbance, this allows the agent to freely explore and thus, as we show in our experiments, enables the agent to learn a policy with a high success rate.
In the second stage, we fine-tune the agent from the previously learned model in the first stage using our new reward function $r'_t$  (Eq.~\ref{eq:reward}). 
In this stage, the agent learns to refine its behavior to better avoid disturbance, without sacrificing performance.
Intuitively, we decompose the disturbance-free objective $r'$ into a task sequence, with the first task being $r$ and the second being $r'$. The first task is easier to learn from scratch, and its goal is closely aligned with the second task.

\section{Experiments}

We evaluate our method on the ArmPointNav task within the manipulation framework ManipulaTHOR~\cite{ehsani2021manipulathor} set within AI2-THOR simulator~\cite{kolve2017ai2}.
Following \cite{ehsani2021manipulathor}, in all experiments we train our agents on the 19 training scenes, tune model hyperparameters (and report the ablation study results) on the 5 validation scenes, and finally report our best-validation-model results on the 5 testing scenes.
Each scene has 720 data points (episodes). Due to space constraints, we report the agent performance when faced with \textit{novel} objects (360 data points) in the main paper and report results on the less-challenging \textit{seen} objects subset in App.~\ref{appendix:additional}. For details on the ManipulaTHOR environment, see App.~\ref{appendix:env}.

We focus on two primary metrics, \textbf{Success Rate (SR)} and \textbf{Success Rate without Disturbance (SRwoD)\footnote{Note that this metric is abbreviated as SRwD in the original paper~\cite{ehsani2021manipulathor}. However we find ``SRwoD'' to be less ambiguous.}}, measuring the original objective and the disturbance-free objective, respectively. 
For each episode, we consider it \textit{successful without disturbance} if it is successful and the \textit{final disturbance distance} $d_{T}^{\mathrm{objects}}$ is less than a threshold (${\approx} 0.01$~m).
We use AllenAct~\cite{weihs2020allenact} as our training framework. We provide more implementation details, such as hyperparameters, in App.~\ref{appendix:training}. All training code and model weights will be made open-source. Qualitative videos can be found in the supplementary materials.

In Sec.~\ref{sec:exp_baselines}, we evaluate several small but critical design decisions that allow us to dramatically improve the SR of the baseline model from Ehsani \etal.
In Sec.~\ref{sec:exp_main}, we describe how our two-stage training approach from Sec.~\ref{sec:curriculum} allows for large gains in SRwoD without sacrificing SR in the final disturbance-free setting.

\subsection{Improved Baseline for ArmPointNav}
\label{sec:exp_baselines}

As described in Sec.~\ref{sec:baseline}, we improve and extend the original architecture for ArmPointNav. Below are the details of our decision choices.
First, inspired by recent works~\cite{wijmans2019dd, wijmans2020train,ye2020auxiliary,mirowski2016learning}, we replace the simple visual encoder with a modified ResNet18~\cite{he2016deep}, use un-normalized advantage estimation in PPO~\cite{schulman2017proximal},\footnote{Please refer to batch-wise normalization in advantage in~\cite{wijmans2020train}.} and add previous actions as an input to the GRU model.
Moreover, we replace the original \textit{unsigned} Cartesian coordinates $(|x|,|y|,|z|)$ in ArmPointNav by polar coordinates $(\rho,\alpha,\beta)$\footnote{where $x=\rho\cos \alpha \sin \beta, y=\rho \sin \alpha \sin \beta,z=\rho \cos \alpha$ following the standard coordinate conversion.} as relative 3D goal coordinates.

Table~\ref{tab:baselines_valid_novel_main} shows how these various design decisions impact the model performance on the \textit{validation} set. All models are trained for 20M simulation steps, following Ehsani \etal.
Combining all the above modifications, we obtain a new baseline model (last row) that greatly outperforms the previous SoTA (row \#1)\footnote{As we use a newer, more physically accurate, version of AI2-THOR~\cite{kolve2017ai2} and smaller batch sizes, the results of our re-implemented baseline are slightly lower than those in the original paper (SR: 61.7\% v.s. 62.1\%; SRwoD: 29.8\% v.s. 32.7\%).} \emph{by 17\% absolute points in SR} in the same training setting.

\begin{table}[t]
    \setlength\tabcolsep{3pt}
    \centering
    \footnotesize
    \begin{tabular}{cccc|cc}
    \toprule
    Visual & Normalized  & Previous & Goal & SR  & SRwoD  \\
    Encoder & Advantage? & Action? & Coordinate & (\%) & (\%)  \\
    \midrule
    CNN & \cmark & \xmark & Cartesian      & 55.8 & 12.3 \\ 
    CNN & \xmark & \xmark & Cartesian      &  59.9 & 16.7  \\
    CNN & \xmark & \cmark & Cartesian      &  52.8 & 12.7 \\
    CNN & \cmark & \cmark & Polar      &  68.9 & 18.3  \\ 
    CNN & \xmark & \xmark & Polar      & 64.2  & 18.3  \\ 
    CNN & \xmark & \cmark & Polar      & 66.5  & 15.3  \\ 
    ResNet & \cmark & \xmark & Cartesian &  60.2 & 13.1  \\ 
    ResNet & \xmark & \xmark & Cartesian &  62.4 & 14.3  \\ 
    ResNet & \xmark & \cmark & Cartesian & 63.6 & 15.9  \\ 
    ResNet & \cmark & \cmark & Polar      &  58.4 & 13.3 \\ 
    ResNet & \xmark & \xmark & Polar      &  67.7 & 14.6  \\ 
     ResNet & \xmark & \cmark & Polar &   \textbf{73.6}  & 18.1   \\
     \bottomrule
    \end{tabular}
    \vspace{2mm}
    \caption{\textbf{Ablating our improved baselines on validation scenes.} Using a ResNet18 visual encoder, un-normalized advantages, polar goal coordinates, and adding previous actions, can significantly increase the performance of the baselines on validation scenes with novel objects. Reported metrics include
    Success Rate (\textbf{SR}) and Success Rate without Disturbance (\textbf{SRwoD}).
    The final improved baseline (last row) outperforms
    the original baseline from~\cite{ehsani2021manipulathor} (first row).}
    \vspace{-2em}
    \label{tab:baselines_valid_novel_main}
\end{table}

\subsection{Results in the Disturbance-Free Setting}
\label{sec:exp_main}

\begin{table*}[t]
    \renewcommand{\arraystretch}{1.05}
    \centering
    \begin{tabular}{ccccc|cc|cc}
    \toprule
    \multirow{2}{*}{Stage} & \multirow{2}{*}{Reward} & \multirow{2}{*}{Initial} & \multirow{2}{*}{Frames} &   \multirow{2}{*}{Aux Task}  & \multicolumn{2}{c|}{SR (\%)} & \multicolumn{2}{c}{SRwoD (\%)}   \\ 
     &  &  & &  & Mean & IQM & Mean & IQM \\
    \midrule
    
     I & $r$ & scratch & 20M &  None  (Original) & 61.7 & - & 29.8 & -  \\ 
     I & $r$ & scratch & 20M &  None (New) &  73.3 & - &  31.7 & - \\
     I & $r$ & scratch & 20M & CPC$|$A~\cite{guo2018neural,ye2020auxiliary} & 74.1 & - & 31.9 & - \\ %
     I & $r$ & scratch & 20M & Inv. Dyn.~\cite{pathak2017curiosity,ye2020auxiliary} & 76.8 & - & 35.0 & - \\  %
     I & $r$ & scratch & 20M &  Disturb (Ours)  &  78.3 & - & 34.0  & -  \\ \hline 
     I & $r$ & scratch & 45M &  None (New) & 82.7 & 82.1 & 35.5 & 35.3  \\ 
     I & $r$ & scratch & 45M & CPC$|$A & 81.4 & 81.6 & 36.2  & 36.9 \\ %
     I & $r$ & scratch & 45M & Inv. Dyn. & 67.8 & 80.9 & 29.4 & 34.9 \\ %
     I & $r$ & scratch & 45M & Disturb  & \textbf{83.5} & \textbf{82.7} & 37.2 & 36.4 \\ \hline
     I & $r'$ & scratch & 45M &  None (New) & 18.0 & 4.8 & 10.5 &  3.0 \\
     I & $r'$ & scratch & 45M & CPC$|$A & 18.2 & 3.6 & 11.1 & 2.1 \\ 
     I & $r'$ & scratch & 45M & Inv. Dyn. & 30.4 & 25.9 & 18.4 & 15.6  \\ 
     I & $r'$ & scratch & 45M & Disturb  & 1.4 & 1.3 & 0.9 & 0.8 \\ \hline
\multicolumn{3}{l}{PPO-Lagrangian~\cite{ray2019benchmarking} ($\lambda_0=1.0$)} & 45M & None (New)  & 30.8 & 36.6 & 15.2 &  18.3 \\
\multicolumn{3}{l}{PPO-Lagrangian ($\lambda_0=15.0$)} & 45M & None (New) & 0.0 & 0.0 & 0.0 & 0.0 \\ \hline
 II & $r'$ & finetune & 20M+25M &  None (New) & 80.1 & 79.9  & 46.5 &   45.9 \\
 II & $r'$ & finetune  & 20M+25M & CPC$|$A  &  79.1 & 78.9 & 46.7 & 46.6  \\ 
 II & $r'$ & finetune  &20M+25M & Inv. Dyn. &  79.6 & 79.8 & 46.9 & \textbf{47.1} \\
 II & $r'$ & finetune  &20M+25M &  Disturb & 81.3 & 81.4 & \textbf{47.1} &  46.6\\ \bottomrule
    \end{tabular}
    \vspace{2mm}
    \caption{\textbf{Main results on testing scenes with novel objects using the large action space $\mathcal A_{\mathrm{large}}$ .} 
    Each method is labeled by its stage in our curriculum (Fig.~\ref{sec:curriculum}), the reward it received ($r$ for original reward; $r'$ for new reward), the weight initialization (from scratch or fine-tuned), number of training frames,  and what auxiliary task it used. For none auxiliary task, ``original'' refers to the original baseline, and ``new'' refers to our improved variant. 
    ``Mean'' column shows the averages over 5 random seeds while ``IQM'' column shows the averages over the 3 seeds with median performance, to reduce the effect of outliers, suggested by~\cite{agarwal2021deep} (see Fig.~\ref{fig:IQM} for plots with confidence intervals).}
    \label{tab:main_test_novel}
    \vspace{-1.5em}
\end{table*}

Before moving to our main results for the disturbance-free setting, we describe one additional design decision: we have enlarged the original action space $\mathcal A_{\mathrm{small}}$ in ArmPointNav into $\mathcal A_{\mathrm{large}}$ so as to include camera and arm rotation actions.
We enlarge the action space in this way as, in  
a qualitative analysis of model failures, we found that agents often appeared to disturb objects in part due to (1) a lack of degrees of freedom in their arm movement and (2) an inability to change their camera's viewing angle to see objects they might disturb.
We also ablate this decision in App.~\ref{appendix:additional} and find that it indeed has an impact on performance.

Table~\ref{tab:main_test_novel} summarizes our main results (using $\mathcal A_{\mathrm{large}}$) when evaluating models on testing scenes with novel objects. We consider 4 training scenarios, corresponding to the 4 main blocks in the table: (\textbf{Block 1}) training from scratch with original objective $r$ for 20M frames; (\textbf{Block 2}) training from scratch with original objective $r$ for 45M frames; (\textbf{Block 3}) training from scratch with new objective $r'$ for 45M frames; and (\textbf{Block 5}) fine-tuning from Block 1 with $r'$ for 25M frames (45M total).

Each row shows the average result over \textbf{5 seeds} for reproducibility. In Block 3 and 5, we use a fixed penalty coefficient $\lambda_{\mathrm{disturb}}=15.0$ for $r'$ after tuning on the validation set (see ablation study in App.~\ref{appendix:additional}). We also run a common safe RL baseline, PPO-Lagrangian~\cite{ray2019benchmarking} (\textbf{Block 4}), with two initial multiplier values $\lambda_0$ (see App.~\ref{sec:lagrangian} for details). 

\noindent\textbf{Auxiliary tasks can improve sample efficiency (Block 1 and 2).} 
For comparison, we consider three different auxiliary tasks: our proposed disturbance prediction task (Sec.~\ref{sec:task}) and two self-supervised tasks, Inverse Dynamics Prediction~\cite{pathak2017curiosity,ye2020auxiliary} and the contrastive CPC$|$A method~\cite{guo2018neural,ye2020auxiliary}. As seen by examining the results in Block 1 (which shows results after 20M training steps), sample efficiency improves when co-training with CPC$|$A and our task. This gain in sample efficiency does not, however, lead to substantial gains in performance after 45M training steps (Block 2).
The agents using our disturbance prediction task perform the best in both regimes, although with a narrower advantage after 45M training steps. 
This demonstrates that an auxiliary task can indeed enhance \textit{sample efficiency}, but not necessarily asymptotic performance.
In our initial experiments we found that combining multiple auxiliary tasks did not meaningfully improve results and so we report using only a single auxiliary task.

\noindent\textbf{Training from scratch learns to stop early with poor success rate (Block 2 and 3).}
Now we move on to the disturbance-free setting. 
The simplest way to train an agent that avoids disturbance is to directly train the policy with the new reward $r'$ (Eq.~\ref{eq:reward}) from scratch (\ie, Block 3). 
However, even when co-training auxiliary tasks, the agents simply fail to learn a reasonable policy in most seeds, with much worse average SR (and also SRwoD) when compared to Block 2 (trained with $r$). 
In fact, 16 out of 20 trials in Block 3 totally fail, with $<$10\% SR. We investigate these failed trials and find that these agents pick up the target objects in 94.1\% of episodes, but only choose to terminate the episode \textit{shortly} after the pick-up within 7.1 steps (recall that the total horizon $T=200$). This means that these agents learn \textbf{a bad local optimum}: pick up the target object to get the pick-up reward bonus, and then immediately terminate the episode to avoid any disturbance penalty. 
Such degradation in SR can be explained as a side effect of disturbance avoidance as suggested in Sec.~\ref{sec:curriculum}.

\noindent\textbf{Two-stage training achieves higher SRwoD without sacrificing SR (Blocks 2, 3, and 5), and much better than PPO-Lagrangian (Block 4), with impressive robustness.}
Our two-stage training curriculum (Block 5) allows agents to avoid degradation in SR, compared to training from scratch (Block 3), while also achieving a much higher SRwoD by $\sim$10\% with similar SR, compared to training on the original reward (Block 2). 
It is also significantly more performant than the safe RL baseline, PPO-Lagrangian, which is sensitive to the initial coefficient.
In the appendix Fig.~\ref{fig:IQM} we show that our approach is also more robust to different seeds than training from scratch or using PPO-Lagrangian. 
Note that all agents were trained for 45M frames. 
Our proposed curriculum is highly effective and robust, easy to implement, and can be used with auxiliary task co-training. Because of these advantages, as embodied-AI tasks begin to take disturbance avoidance more seriously in an effort to enable real-world deployment, we expect that our approach's simplicity and robustness will be its great advantages: any researcher in embodied AI can easily leverage our training approach to enable disturbance avoidance in their models.

\noindent\textbf{Emergence of temporary displacement.}
As noted previously, our reward structure $r'$ allows agents to disturb objects, so long as they eventually move those objects back to their original positions (approximately). Perhaps surprisingly, we find that this behavior emerges in testing scenes: 
in a quantitative analysis, we find that the agents in Block 5 (Table~\ref{tab:main_test_novel}) learn to temporarily move the other objects (and then recover their positions) in ${\approx}$5\% of episodes.

\begin{figure}[h]
    \centering
    \includegraphics[width=\linewidth]{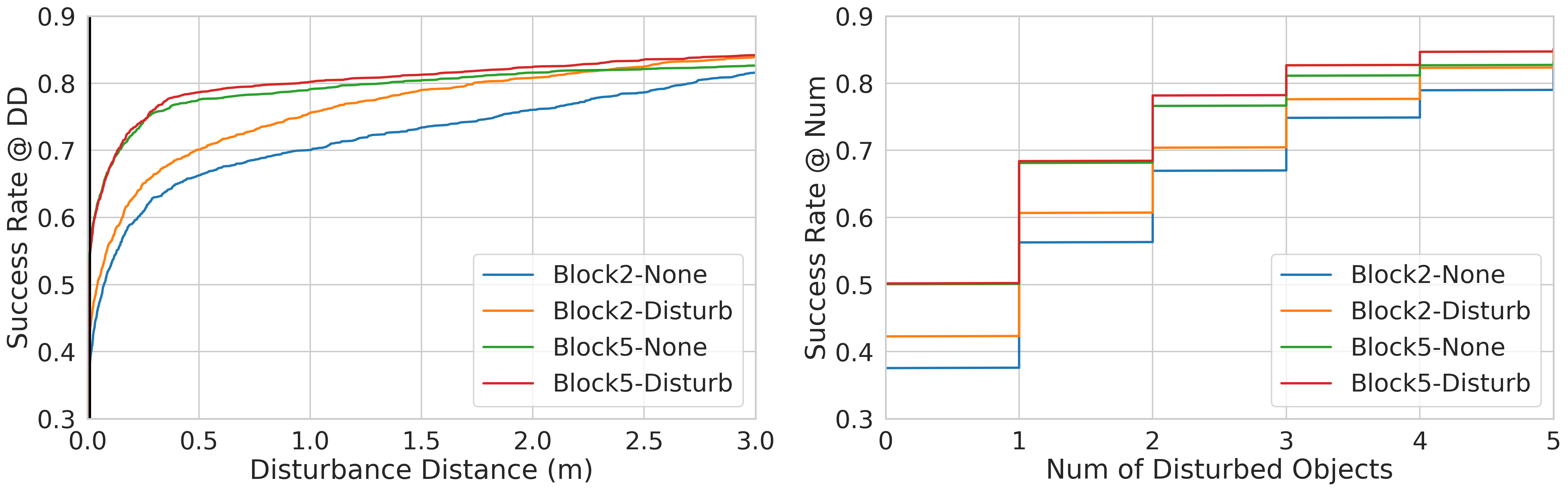}
    \vspace{-1em}
    \caption{\footnotesize \textbf{Success rate without disturbance curves.} Each method is labeled by its training setting (Block 2 or 5) and the auxiliary task it uses, all trained with 45M frames.
    The x-axis in the left figure is the disturbance distance ($d_{T}^{\mathrm{objects}}$, $\mathrm{DD}$ in short), and y-axis is the \% of episodes that are \emph{both} successful and have $\mathrm{DD}$ lower than the x value.
    The intersection of the vertical line at $\mathrm{DD}=0.01$ and each curve is ${\approx}$SRwoD. Similarly, the right figure uses the number of disturbed objects as its x-axis.}
    \vspace{-1.5em}
    \label{fig:curve}
\end{figure}

\noindent\textbf{Success rate at various disturbance thresholds.}
To show the detailed results on the relationship between disturbance distance and success rate, we plot the success without disturbance curves in Fig.~\ref{fig:curve} for our best performing models. We can see that Block 5 agents indeed outperform Block 2 agents in all the disturbance distance thresholds less than 2.5 meters, reaching around 80\% success rate when the disturbance distance threshold is less than 1 meter or at most 2 objects are disturbed.

\begin{figure}[h!]
    \centering
    \includegraphics[width=\linewidth]{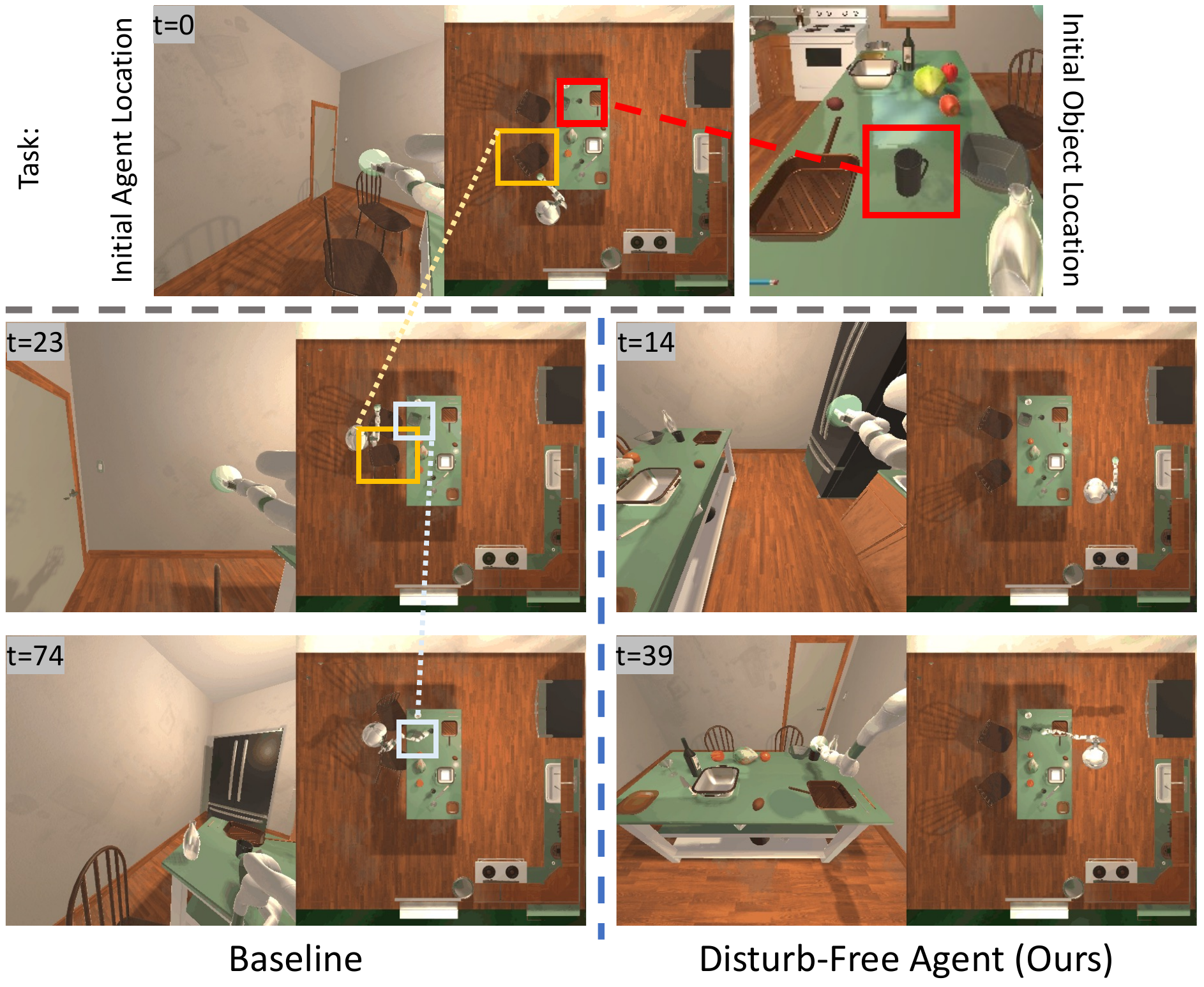}
    \caption{\footnotesize \textbf{Qualitative improvements.} The task is to pick up the \textcolor{red}{mug} on the table starting from the stove. 
    Our disturb-free agent  observes the chairs and plans accordingly to avoid disturbing the scene. It successfully picks up the object. 
    In contrast, our baseline agent disturbs the \textcolor{green}{chairs} with its body and the \textcolor{cyan}{bowl} on the table with its arm. The steps it spent colliding with objects prevent it from picking up the mug before the episode finishes.}
    \vspace{-2em}
    \label{fig:qualitative}
\end{figure}

\noindent\textbf{Qualitative results.}
Fig.~\ref{fig:qualitative} shows an example of how our disturbance-free approach can achieve better performance on the task of ArmPointNav. See our supplementary materials for more qualitative results.

\section{Conclusion}

This paper highlights the importance of disturbance (collision) avoidance to embodied AI, with a focus on visual mobile manipulation. 
We first formalize the objective of disturbance avoidance for RL agents, and then provide extensive evidence that our two-stage training curriculum is much more effective and robust than training from scratch on that objective, leading to state-of-the-art performance on success rates without disturbance in ArmPointNav task. 
Moreover, we propose a new auxiliary task of disturbance prediction to improve sample efficiency. 
Although we evaluate our method only in ArmPointNav task due to the scarcity of benchmarks,  
we believe that all the components of our method, including the disturbance-free objective, the two-stage training curriculum, and the new auxiliary task, are general, and could be applied to other tasks to accelerate safe deployment of robots in the real world.

\subsection*{Acknowledgement}
We thank Aniruddha Kembhavi and Roozbeh Mottaghi for their insightful feedback on the early draft of the paper.

{\small
\bibliographystyle{ieee_fullname}
\bibliography{citation}
}

\clearpage
\appendix

\section*{Appendices for \emph{Towards Disturbance-Free Visual Mobile Manipulation}}

\section{Environment Details}
\label{appendix:env}
Our experimented environment directly follows the visual mobile manipulation task, ArmPointNav, in the ManipulaTHOR framework~\cite{ehsani2021manipulathor}. The details of the task and environment can be seen in the ManipulaTHOR paper. Here, we give a high-level overview of this task. 

ArmPointNav has a dataset called APND that stores the configurations of each episode. APND consists of 29 kitchen scenes in AI2-THOR~\cite{kolve2017ai2} that have more than 150 object categories. 
Among the scenes, 19 of them are used for training, 5 of them for validation, and 5 for testing. 
There are 12 pickupable categories, and 6 of them (Apple, Bread, Tomato, Lettuce, Pot, Mug) are used for training (\ie, seen objects), and the others (Potato, SoapBottle, Pan, Egg, Spatula, Cup) are used for testing and validation (\ie, novel objects). 

Each episode has a configuration that specifies the initial and target positions of the target object and the initial position of the agent. The goal for the agent is to first navigate towards the target object, pick it up with a magnet, and then navigate towards the target location to release the object. 

Mathematically, ArmPointNav can be formulated as a POMDP $(\mathcal S, \mathcal O, \mathcal A, R, T, \gamma, P, O)$. Each state  $s\in \mathcal S$ includes the 3D positions of the robot, the goal, and all the obstacles in the room. 
Each observation $o\in \mathcal O$ includes a depth map ($224\times 224$ single-channel image) and a distance coordinate to the goal (3 dimensions). The goal switches from the initial position of the target object to the desired position of the target object once the agent picks it up. The action space ($\mathcal A_{\mathrm{large}}$) has 21 actions including: (1) navigation (move ahead, rotate), (2) manipulation (move the arm and gripper), (3) camera rotation, and (4) pick up and done. The details of the action space can be seen in Fig.~\ref{fig:heatmap}. The reward function $R$ is defined in Eq.~\ref{eq:reward} and Eq.~\ref{eq:disturb_free_obj}. The time horizon $T = 200$ steps, and the discount factor $\gamma = 0.99$. The transition $P(s_{t+1} \mid s_t,a_t)$ determines how the robot and obstacles move in 3D coordinates, and the emission $O(o_t \mid s_t)$ determines the rendering of egocentric vision to the robot.

The task is a POMDP rather than an MDP because the robot cannot observe the ground-truth positions of the obstacles nearby, which are crucial for optimal control with a Markovian policy. It can only use the historic information of egocentric depth maps to infer them.

\section{Experiment Details}
\label{appendix:training}
We train our agents using the AllenAct framework~\cite{weihs2020allenact}. 
All the experiments including baselines and compared self-supervised auxiliary tasks share most training hyperparameters. Each experiment uses 19 processes (each sampling rollouts on one training scene) and trains for 45M frames (for two-stage curriculum: pre-training for 20M frames, and then fine-tuning for 25M frames).

We use the DD-PPO algorithm~\cite{wijmans2019dd} with default configuration. The model architecture (Fig.~\ref{fig:arch}) uses a modified ResNet18 with group normalization~\cite{wu2018group} following DD-PPO as the visual encoder, an embedding layer into 32-dim for goal coordinates, an embedding layer into 16-dim for previous actions, then a GRU with a hidden size of 512, and finally linear actor and critic heads.

For the self-supervised auxiliary tasks CPC$|$A and Inverse Dynamics, we directly follow the implementation of Ye~\etal~\cite{ye2020auxiliary}\footnote{\url{https://github.com/joel99/habitat-pointnav-aux}}. 

For our disturbance prediction auxiliary task, we use a 2-layer MLP of hidden size 128 to predict the disturbance distance signals $\in [0,1]^{|\mathcal A_{\mathrm{large}}|}$, for all the actions $\in \mathcal A_{\mathrm{large}}$ (similarly to Deep Q-Network~\cite{mnih2013playing}), given the current belief $\in \mathbb R^{512}$.
The auxiliary task uses the Focal loss~\cite{lin2017focal} with $\gamma=2.0$ and $\alpha=0.5$. 
The overall objective is a weighted sum of the RL loss and the auxiliary task loss, with a fixed weight of $0.1$ on the auxiliary task, following Ye~\etal.

\section{PPO-Lagrangian Details}
\label{sec:lagrangian}

PPO-Lagrangian is a common baseline from the safe RL literature~\cite{ray2019benchmarking} which aims to solve constrained MDPs~\cite{altman1999constrained}. 
In our paper, the original objective is the ArmPointNav task with original reward, 
\begin{equation}
J(\pi):= \E{\tau\sim \pi}{\sum_{t=1}^T \gamma^t r_t}\ ,
\end{equation}
and the constraint is on the total disturbance distance, which we want to be non-positive:
\begin{equation}
J_C(\pi):= \E{\tau\sim \pi}{\sum_{t=1}^T \gamma^t (d_{t}^{\mathrm{objects}} - d_{t-1}^{\mathrm{objects}})}{\le}0 \ .
\end{equation}

The corresponding Lagrangian is:
\begin{equation}
\min_{\lambda \ge 0} \max_\pi J(\pi) - \lambda J_C(\pi) \ ,
\end{equation}
where $\lambda$ is the Lagrangian multiplier.  

The Lagrangian method alternatively update the policy $\pi$ and the Lagrangian multiplier $\lambda$: 
\begin{itemize}
\item Given current Lagrangian multiplier $\lambda_k$  and the learning rate $\eta_\pi$, the policy $\pi_k$ is updated by
    \begin{equation}
    \pi_{k+1}{\gets} \pi_k + \eta_\pi \nabla_\pi (J(\pi_k) - \lambda_k J_C(\pi_k)) \ .
    \end{equation}
\item Given the current policy $\pi_{k+1}$ and the learning rate $\eta_\lambda$, the Lagrangian multiplier $\lambda_k$ is updated by
    \begin{equation}
    \lambda_{k+1}{\gets}(\lambda_k + \eta_\lambda J_C(\pi_{k+1}))_+ \ .
    \end{equation}

\end{itemize}
The initial value of Lagrangian multiplier $\lambda_0$ is set by the user, and is shown to be crucial to performance~\cite{achiam2017constrained}. 

Applying the Lagrangian method to the PPO algorithm, we obtain the PPO-Lagrangian method. We use the same hyperparameters in (DD-)PPO as that in our method.

\section{Additional Results}
\label{appendix:additional}

\subsection{Reliable Evaluation Plots}
\label{app:IQM}
\begin{figure*}[h]
    \centering
    \includegraphics[width=0.8\linewidth]{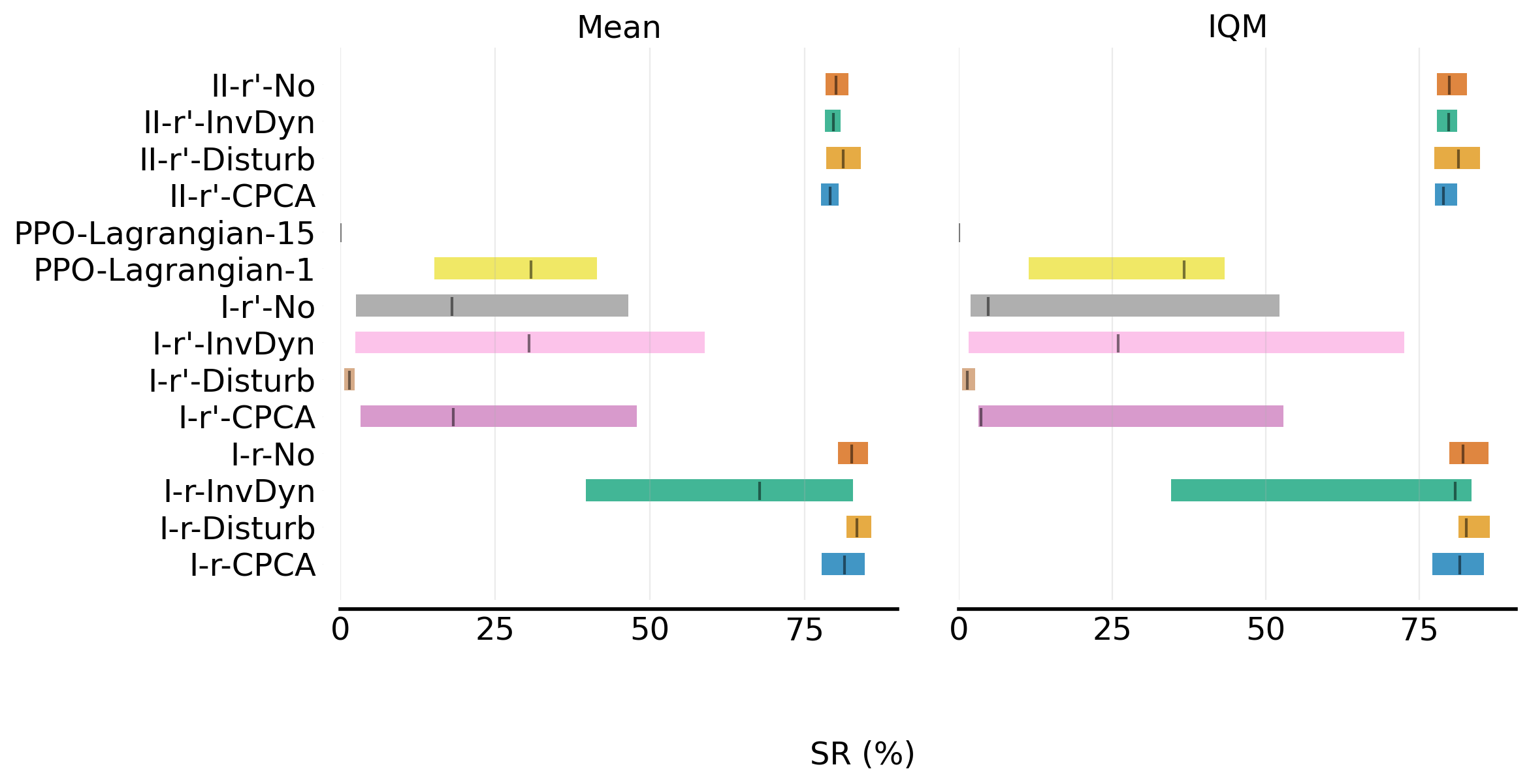}
    \includegraphics[width=0.8\linewidth]{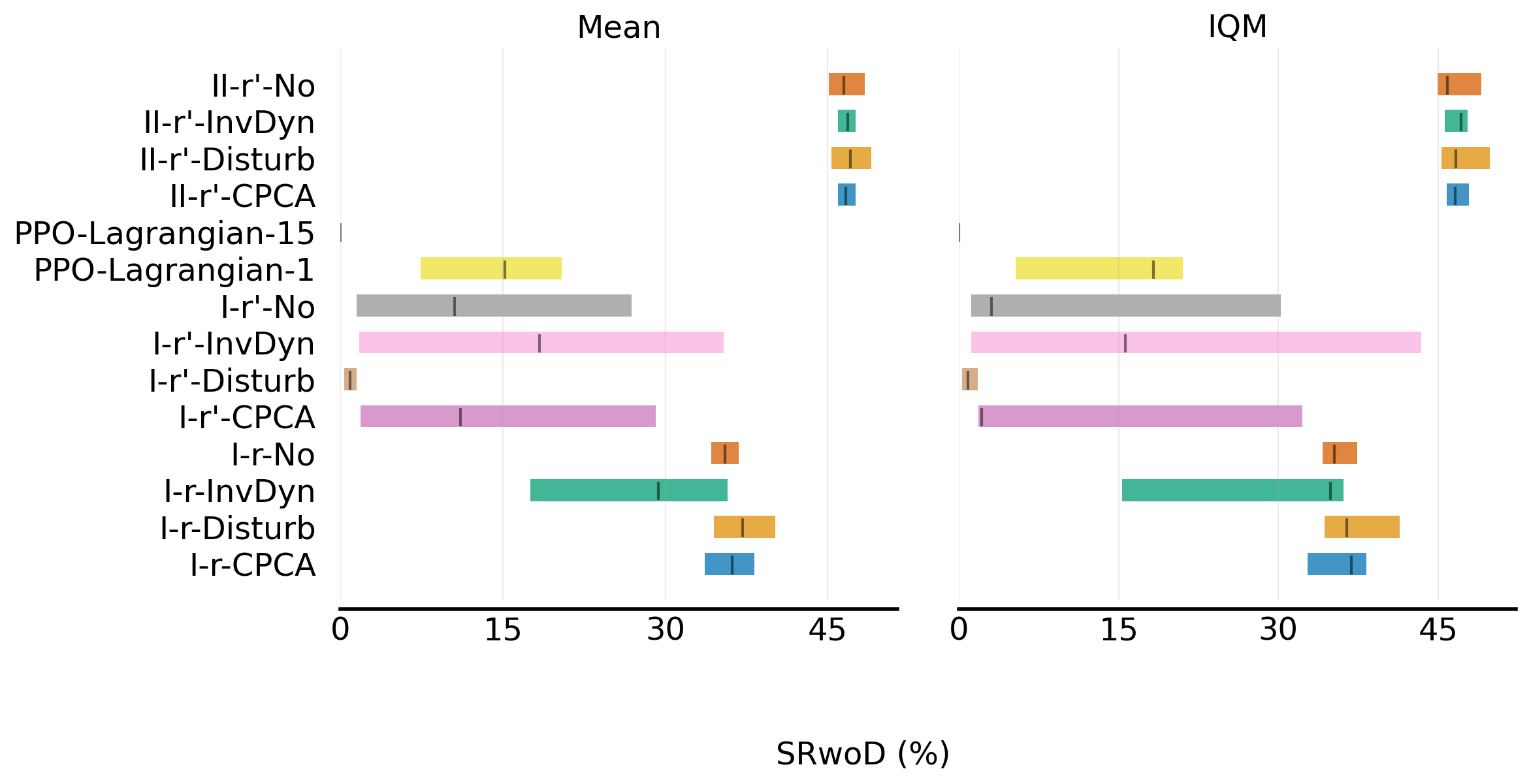}
    \caption{\textbf{Mean and IQM with 95\% (stratified bootstrap) confidence intervals} on \textbf{SR} (top figure) and \textbf{SRwoD} (bottom figure), following the rliable library~\cite{agarwal2021deep}. We denote each method by its stage (I, II), reward ($r$,$r'$), and auxiliary task. 1 and 15 in PPO-Langragian stand for the initial value of the multiplier. All methods are trained for 45M frames.}
    \label{fig:IQM}
\end{figure*}

We follow the rliable library\footnote{\url{https://github.com/google-research/rliable}} to evaluate our method and baselines. Fig.~\ref{fig:IQM} shows the mean and IQM of SR and SRwoD metrics (reported in Table~\ref{tab:main_test_novel}), and also their 95\% confidence intervals (CIs).  

We find that fine-tuning stage (stage II) is much more robust to seeds with narrower CIs, than trained-from-scratch stage (stage I), and also PPO-Lagrangian. This suggests the robustness and reliability of our method.

\subsection{Main Results on Testing Scenes with Seen Objects}

Table~\ref{tab:main_test_seen} shows the main results on testing scenes with seen objects (recall that Table~\ref{tab:main_test_novel} is on testing scenes with novel objects). 
Generally speaking, the trend in Table~\ref{tab:main_test_novel} holds in Table~\ref{tab:main_test_seen}:  
\begin{itemize}
\setlength\itemsep{1mm}
    \item Auxiliary tasks can improve sample efficiency (Block 1 and 2). 
    \item Training from scratch learns to stop early with poor success rate (Block 2 and 3).
    \item Two-stage training achieves higher SRwoD without sacrificing SR (Blocks 2, 3, and 5).
    \item Two-stage training outperforms the safe RL baseline (Block 4 and 5).
\end{itemize}

Comparing Table~\ref{tab:main_test_seen} to Table~\ref{tab:main_test_novel}, we find that the SR in Block 1, 3, and 5 increases by $\approx$ 1-2\% for all the auxiliary tasks, and SRwoD in Block 5 increases by $\approx$ 2\%. This is understandable because Table~\ref{tab:main_test_seen} are evaluated on seen objects.

\subsection{Effect of Disturbance Penalty Coefficient}

The disturbance-free objective in Eq.~\ref{eq:disturb_free_obj} is sensitive to the disturbance penalty coefficient $\lambda_{\mathrm{disturb}}$. Ideally, the coefficient should be large enough to enforce a hard constraint on disturbance. But a too-large coefficient may hinder the agent from reaching the goal with a very small disturbance distance, thus affecting the success rate. To balance between SR and SRwoD, one has to tune the coefficient on a validation set.

\begin{figure}[h]
    \centering
    \includegraphics[width=0.8\linewidth]{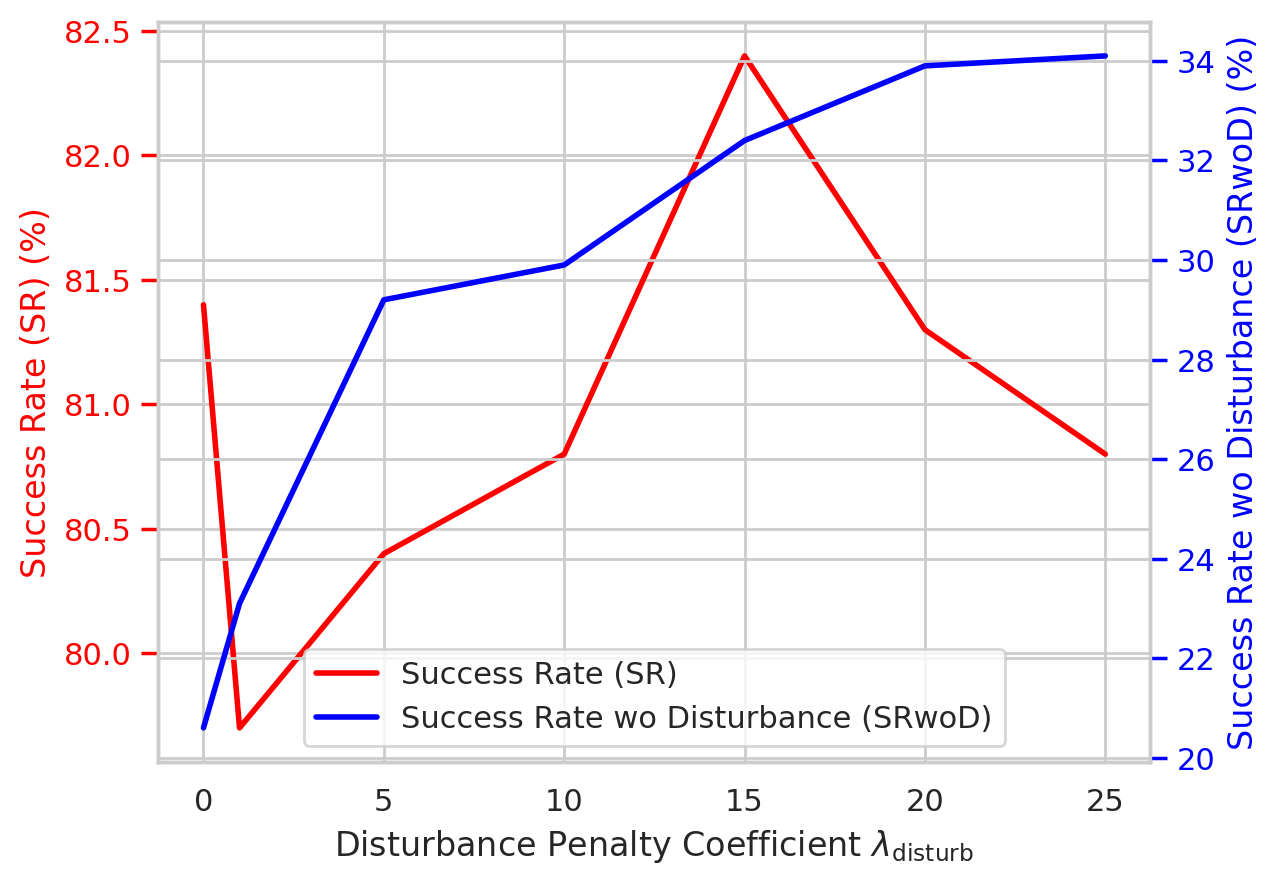}
    \caption{\textbf{The effect of disturbance penalty coefficient $\lambda_{\mathrm{disturb}}$ on validation scenes.} The curves show the final results of fine-tuning with disturbance prediction task after 25M steps with different $\lambda_{\mathrm{disturb}}$.}
    \label{fig:fine-tuning}
\end{figure}

Fig.~\ref{fig:fine-tuning} shows the performance of the fine-tuned models with disturbance prediction auxiliary tasks (Block 5), with different penalty coefficients $\lambda_{\mathrm{disturb}}$.
SRwoD monotonically increases from $79.5\%$ to $82.5\%$ with a larger disturbance penalty coefficient, which shows the effectiveness of our method. 
We finally choose $\lambda_{\mathrm{disturb}}=15.0$ because it balances SR and SRwoD best. Surprisingly, disturbance avoidance ($\lambda_{\mathrm{disturb}}=15.0$) can even help success rate, compared to the model with original reward (\ie, $\lambda_{\mathrm{disturb}}=0.0$). Note that these experiments are only ablations over the validation set and we only calculate the final performance of our model on the test set when using $\lambda_{\mathrm{disturb}}=15.0$.

\subsection{Is the \emph{Large} Action Space Necessary for Disturbance Avoidance?} 

As described at the beginning of Sec.~\ref{sec:exp_main}, we found, qualitatively, that the original action space hinders the ability of agents to perform disturbance-free tasks. \Eg, the agent may not even be able to see the disturbance it causes, as it cannot look down.
Table~\ref{tab:action_space_main} shows our ablation study on the action space on the \textit{validation} set. Interestingly, under the original objective (Block 1; reward $r$) in ArmPointNav and without auxiliary tasks, the performance of our baseline drops when switching from the small action space to a large one (Rows 1-2 and 5-6). But with our disturbance prediction task (Row 3 and 7), the performance increases by 2.4\%. 

However, the large action space helps with achieving a disturbance-free agent. Our best model (Block 5) performs better with the \emph{large} action space when trained with the new disturbance-free objective (Row 4 and 8).

Moreover, by qualitatively examining the actions taken by well-trained agents (Block 5), we find that, in the large action space setting, agents almost always take the \texttt{LookDown} $\in \mathcal A_{\mathrm{large}}\setminus\mathcal A_{\mathrm{small}}$ as their first action allowing them a better view of the impact of their actions 
(see App.~\ref{appendix:action}). Thus, quantitatively and qualitatively, we find that the added actions indeed help avoid disturbance.

\subsection{Agent Action Distribution}
\label{appendix:action}

\begin{figure*}[h]
    \centering
    \includegraphics[width=\linewidth]{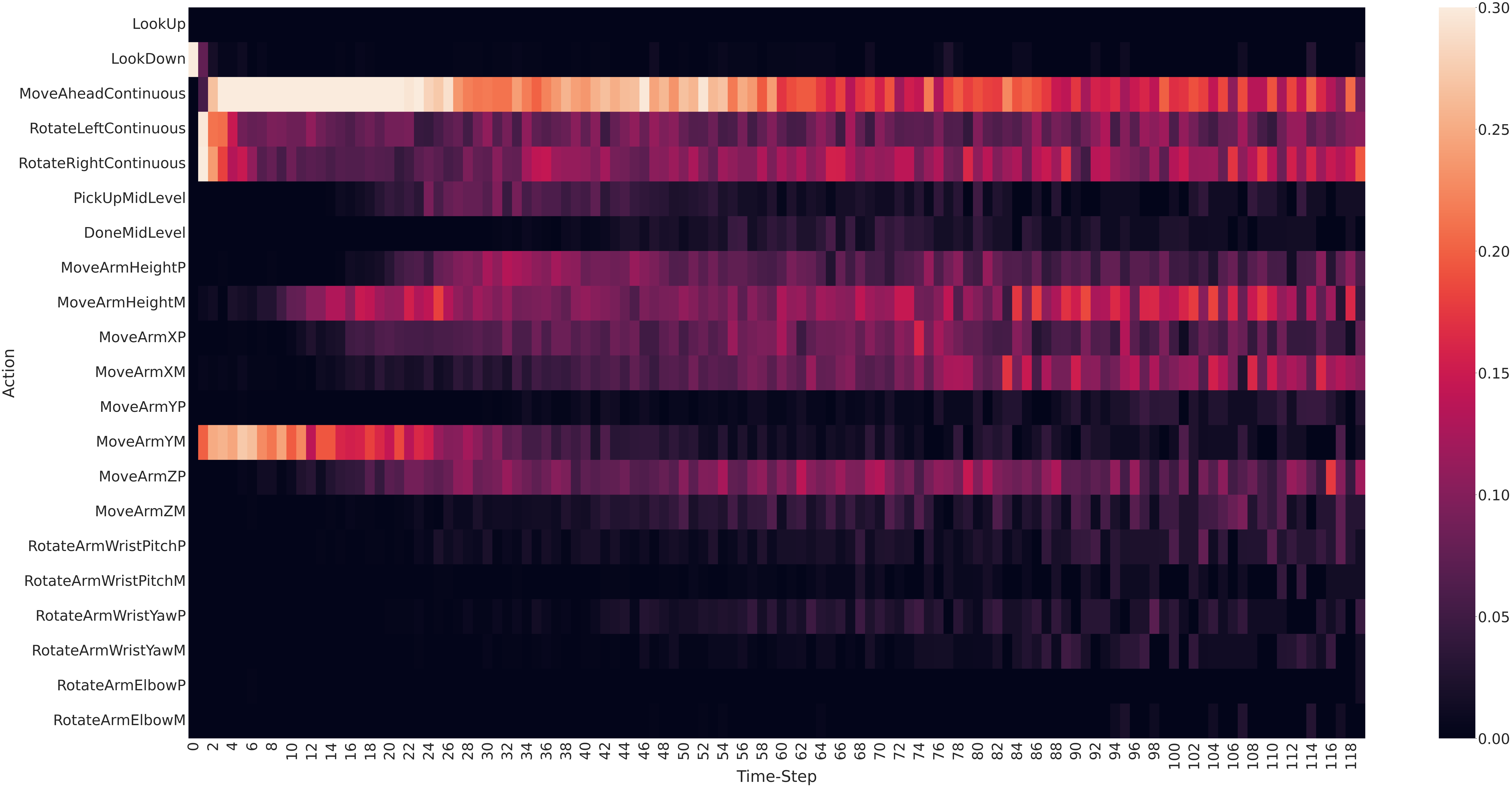}
    \caption{\textbf{Heatmap of the action distribution of our best agent over time.} The y-axis lists all possible actions in the large action space ($\mathcal A_{\mathrm{large}}$). The x-axis shows the time-step from 0 to 120. The sum of all the cells in each column (time-step) is $1.0$. We clip the cell value to $0.3$ for better visualization.}
    \label{fig:heatmap}
\end{figure*}

To better understand how the agent behaves across time, we plot the heatmap of the action distribution of our best agent (the last row of Table~\ref{tab:main_test_novel}) in Fig.~\ref{fig:heatmap}. We show the action distribution for each time-step averaged across episodes. This visualization gives us several insights into the agent's behavior.
Firstly, \texttt{LookDown} is almost always taken at the first time-step to enable the agent to have a better perspective, this justifies the necessity of using the large action space for the disturbance-free objective. 
Secondly, we can clearly see that there are two phases during an episode. The first phase is to move to pick up the target object (roughly from time-step 0 to 45), where the agent first moves ahead (\texttt{MoveAheadContinuous}) and rotates (\texttt{RotateLeft/RightContinuous}) to navigate, then the agent moves the arm and gripper downwards  (\texttt{MoveArmHeightM}, \texttt{MoveArmYM}), and finally the agent picks up (\texttt{PickUpMidLevel}) the target object. 
The second phase focuses on taking the object to the target location (roughly from time-step 40 to 100). Similar to the first phase, the agent first moves ahead and rotates to navigate, and then moves the arm and gripper downwards. But the agent also demonstrates delicate arm behavior during the second phase, such as moving the gripper in XY plane (\texttt{MoveArmX/Y*}) and rotating the arm wrist (\texttt{RotateArmWrist*}), to reduce disturbance to the other objects when placing the target object.

\section{Code and Videos}
Our code is available at \url{https://github.com/allenai/disturb-free}.
The videos of our method and compared methods can be accessed at \url{https://sites.google.com/view/disturb-free}.

\begin{table*}[t]
    \renewcommand{\arraystretch}{1.1}
    \centering
    \begin{tabular}{ccccc|cc}
    \toprule
    Stage & Reward & Initial & Frames & Aux Task  & SR (\%) & SRwoD (\%)   \\ \midrule
     I & $r$ & scratch & 20M &  None  (Original) & 66.3 & 32.1   \\ 
     I & $r$ & scratch & 20M &  None (New) & 74.7 & 32.2  \\
     I & $r$ & scratch & 20M & CPC$|$A~\cite{guo2018neural,ye2020auxiliary} & 76.5 & 31.6 \\ 
     I & $r$ & scratch & 20M & Inv. Dyn.~\cite{pathak2017curiosity,ye2020auxiliary} & 78.3 & 34.3 \\ 
     I & $r$ & scratch & 20M &  Disturb (Ours)  &  79.6 & 34.4   \\ \hline 
     I & $r$ & scratch & 45M &  None (New) &  \textbf{84.1} & 35.8 \\
     I & $r$ & scratch & 45M & CPC$|$A & 82.4 & 36.7 \\ %
     I & $r$ & scratch & 45M & Inv. Dyn. & 69.3 & 28.7 \\ %
     I & $r$ & scratch & 45M & Disturb  & 84.0 & 36.7 \\ \hline
     I & $r'$ & scratch & 45M &  None (New) & 18.4 & 10.1    \\
     I & $r'$ & scratch & 45M & CPC$|$A &  17.9 & 10.4 \\ 
     I & $r'$ & scratch & 45M & Inv. Dyn. &  31.2 & 18.1 \\ 
     I & $r'$ & scratch & 45M & Disturb  & 1.4 & 0.6  \\ \hline
\multicolumn{3}{l}{PPO-Lagrangian~\cite{ray2019benchmarking} ($\lambda_0=1.0$)} & 45M & None (New)  & 33.3  & 15.6 \\
\multicolumn{3}{l}{PPO-Lagrangian ($\lambda_0=15.0$)} & 45M & None (New) & 0.0 & 0.0 \\ \hline
 II & $r'$ & finetune & 20M+25M &  None (New) & 80.8 & 49.0  \\
 II & $r'$ & finetune  & 20M+25M & CPC$|$A  &  79.6 & 47.5  \\ 
 II & $r'$ & finetune  &20M+25M & Inv. Dyn. & 80.9 & 48.8 \\
 II & $r'$ & finetune  &20M+25M &  Disturb & 81.7 & \textbf{49.4}  \\ \bottomrule
    \end{tabular}
    \vspace{2mm}
    \caption{\textbf{Main results on testing scenes with seen objects using the large action space $\mathcal A_{\mathrm{large}}$ .} 
    Each method is labeled by its stage in our curriculum (Fig.~\ref{sec:curriculum}), the reward it received ($r$ for original reward; $r'$ for new reward), the weight initialization (from scratch or fine-tuned), number of training frames,  and what auxiliary task it used. For none auxiliary task, ``original'' refers to the original baseline, and ``new'' refers to our improved variant. Results are averaged over 5 random seeds.}
    \label{tab:main_test_seen}
\end{table*}

\begin{table*}[h]
     \renewcommand{\arraystretch}{1.1}
    \centering
    \begin{tabular}{ccccc|cc}
    \toprule
   Row & Action Space &  Reward & Initial & Aux Task & SR (\%)  & SRwoD  (\%)   \\ \midrule
    1 & $\mathcal A_{\mathrm{small}}$ &  $r$ & scratch & None (Original) &  55.8 & 12.3  \\ 
    2 & $\mathcal A_{\mathrm{small}}$ &  $r$ & scratch & None (New) & 73.7 & 18.1 \\ 
    3 & $\mathcal A_{\mathrm{small}}$ &  $r$ & scratch & Disturb & 70.0 & 16.3  \\   
 4 & $\mathcal A_{\mathrm{small}}$&   $r'$ & finetune  &Disturb & \textbf{74.2} &  \textbf{26.2}  \\ \hline 
    5 & $\mathcal A_{\mathrm{large}}$ &  $r$ & scratch & None  (Original) & 56.4 & 11.9  \\ 
    6 & $\mathcal A_{\mathrm{large}}$ &  $r$ & scratch & None (New) &  66.8 & 16.9  \\ 
    7 &$\mathcal A_{\mathrm{large}}$ &  $r$ & scratch & Disturb &  72.9 & 17.0  \\ 
8 & $\mathcal A_{\mathrm{large}}$&   $r'$ & finetune  &Disturb & \textbf{78.5} & \textbf{29.7}  \\  
\bottomrule
    \end{tabular}
    \vspace{2mm}
    \caption{\textbf{\emph{Large} action space can increase the performance on disturbance avoidance.} $\mathcal A_{\mathrm{small}}$ stands for the original action space, while $\mathcal A_{\mathrm{large}}$ represents the augmented action space this paper adopted. The counterparts (Row 1 \& 5, 2 \& 6, 3 \& 7, 4 \& 8) are trained with same setting except for the action space. Results are on validation set.}
    \label{tab:action_space_main}
\end{table*}

\end{document}